\documentclass[a4paper,fleqn]{article}
\usepackage{INTERSPEECH2021,amssymb,amsmath,epsfig}
\usepackage{xspace}
\usepackage{srcltx}
\usepackage{caption}
\usepackage{subcaption}
\usepackage[export]{adjustbox}
\usepackage{multirow}
\usepackage{amsmath}
\usepackage{url}
\usepackage{hyperref}
\ninept
\def\reg{{\rm\ooalign{\hfil
     \raise.07ex\hbox{\scriptsize R}\hfil\crcr\mathhexbox20D}}}

\usepackage{color}
\usepackage[dvipsnames]{xcolor}

\newcommand{\CMT}[1]{{}}
\usepackage{soul}

\def\vx{\mathbf{x}}

\def\pause{\texttt{<pause>} }

\def\eos{\texttt{</s>} }

\def\reg{{\rm\ooalign{\hfil
     \raise.07ex\hbox{\scriptsize R}\hfil\crcr\mathhexbox20D}}}

\def\vx{\mathbf{x}}

\ninept
\begin{document}
\title{Turn-Taking Prediction for Natural Conversational Speech}
\name{Shuo-yiin Chang, Bo Li, Tara N. Sainath, Chao Zhang, Trevor Strohman, Qiao Liang, Yanzhang He}
\address{Google Inc., U.S.A}
\email{\{shuoyiin,boboli,tsainath,chaoz,strohman,wildstone,yanzhanghe\}@google.com}

\maketitle
\begin{abstract}

While a streaming voice assistant system has been used in many applications, this system typically focuses on unnatural, one-shot interactions assuming input from a single voice query without hesitation or disfluency. However, a common conversational utterance often involves multiple queries with turn-taking, in addition to disfluencies. These disfluencies include pausing to think, hesitations, word lengthening, filled pauses and repeated phrases. This makes doing speech recognition with conversational speech, including one with multiple queries, a challenging task. To better model the conversational interaction, it is critical to discriminate disfluencies and end of query in order to allow the user to hold the floor for disfluencies while having the system respond as quickly as possible when the user has finished speaking.  In this paper, we present a turn-taking predictor built on top of the end-to-end (E2E) speech recognizer. Our best system is obtained by jointly optimizing for ASR task and detecting when the user is paused to think or finished speaking. 
%Our best system is obtained by sharing the encoder and the prediction network from the E2E speech recognizer while adapting the joint layers to optimizes both the recognition and turn-taking detection. 
The proposed approach demonstrates over 97\% recall rate and 85\% precision rate on predicting true turn-taking with only 100 ms latency on a test set designed with 4 types of disfluencies inserted in conversational utterances.
\\
\end{abstract}
%  To better understand how the turn-taking decision is made, we investigate the models relied on acoustic features or semantic history alone.

\section{Introduction}
\label{sec:intro}

Streaming speech recognition systems have been widely used in many voice interaction applications e.g. voice assistant and dialog systems. To achieve a human-level conversational experience, it is essential to learn interaction patterns that resemble human conversational turn-taking. One of the problems is determining when the user has finished speaking, which is typically referred as endpointing \cite{ibm2015, chang2019joint, chang2019unified, chang2017endpoint, li2020towards, amazon2018}. A typical endpointer model makes a series of binary decisions: to wait further for more speech, or to stop listening. While it has been used in many voice systems, these systems often assume "fluent" one-shot voice commands or search queries, where users know exactly what they want to say beforehand. However, a natural conversational utterance commonly involves disfluencies including pauses to think, hesitations, word lengthening, filled pauses (e.g., 'uh', 'um'), and repeated phrases. The disfluencies introduce long pauses in the utterances which could easily cause ambiguity to the E2E model that the user is done speaking. Thus, modeling disfluencies is critical to ensure natural conversational interaction. 

\begin{figure}[h!]
\centering
\hspace{-0.05in}
    \includegraphics[scale=0.27]{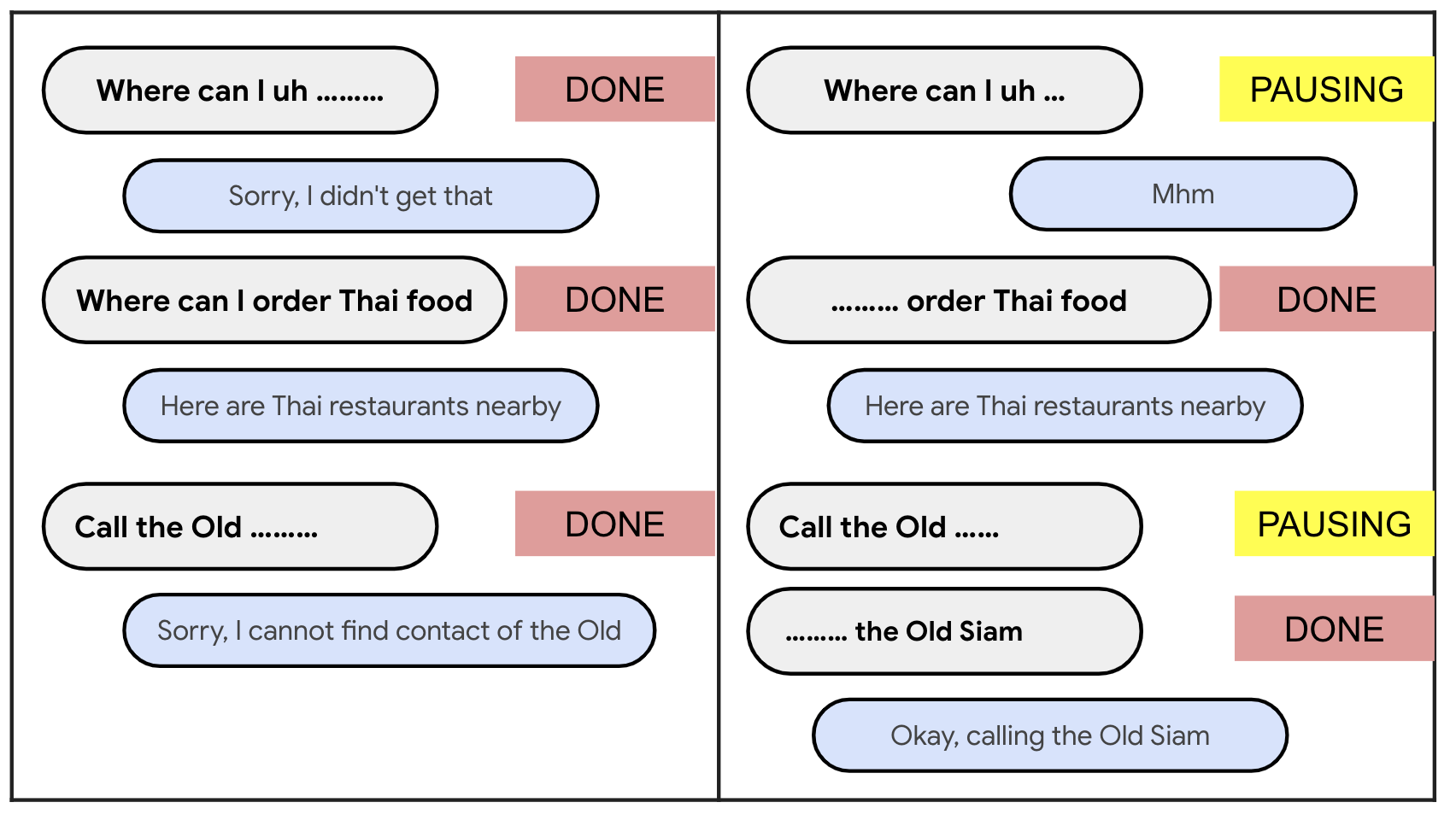}
    \vspace{-0.1in}
    \caption{Conventional endpointer (left) and conversational turn taking model (right)}
    \vspace{-0.1in}
    \label{fig:natcon_example}
\end{figure}

To better model the natural conversational interaction, we propose to build a turn-taking model detecting both when the user pauses to think or when they finish speaking under disfluent natural conversation. It is desirable to respond to users immediately when users have done speaking while allowing users to hold the floor if users are pausing to think.
%\TS{this sentence is not clear to me}.
%Figure \ref{fig:natcon_example} demonstrates that discriminating pausing and finishing could significant improve the conversational interaction experience. 
As illustrated Figure \ref{fig:natcon_example}, the conventional endpointer does not allow users to speak with long pauses or filler words, but instead cuts off the user with an inappropriate response. On the other hand, the proposed turn-taking model generates extra cues of pausing that allows the user speak with disfluencies to achieve a conversational interaction experience.

%\TS{you should describe fig 1 in the text a bit more.} 
%\TS{somewhere you need 1-2 paras on related work. turn-taking is not new. what is novel/different about your work?}

Past studies about turn-taking models \cite{icml18, Osakais17, EOTis17} exploit both acoustic and language model information to predict turn-taking related classes including wait, done speaking or backchanneling e.g. “uh-huh”, “yes”. 
In \cite{kyoto18, filler19}, fillers detection is also explored to assist the turn-taking model.
The acoustic features e.g., prosody, are investigated in \cite{Osakais17, kyoto18} to detect the pauses or a pitch reset at strong phrase boundaries.
%The language model is typically based on the word embedding obtained by the ASR top hypothesis. 

The recent development of end-to-end (E2E) models \cite{li2020comparison,Ryan19,CC18,JinyuLi2019,Zeyer2020}  has already shown that having one neural network to do acoustic, pronunciation and language modeling is far better then a modular-based conventional ASR model\cite{Golan16}. Furthermore, we have seen that folding additional detectors into the E2E model, for example the endpointer \cite{chang2017endpoint, li2020towards} is far better then having separate modules. Building on this, we propose to build an E2E model that incorporates the turn-taking detector into the E2E model that already folds different components of the speech recognition pipeline into one neural network. This is unlike the previous systems that build an external turn-taking model. The proposed E2E turn-taking detector sees both acoustic representations and intonation patterns from the encoder, as well as grammatically unfinished or finished sentences by decoder, to help aid its performance.

Our best system is obtained by sharing the encoder and the prediction network from the E2E speech recognizer while adapting the joint layers to optimize both the recognition and turn-taking detection. The proposed E2E approach provides 97\% recall rate and 85\% precision rate on predicting true turn-taking with only 100 ms latency over a test set including 4 types of disfluencies inserted in conversational utterances. The experiments also investigates the acoustic based approach, text based approach and the E2E model on pausing and finishing detection. 

%\TS{you should have a para describing high level the modeling approach and then another para on the results.}
\section{Training Data Annotation} \label{sec:annonation}

\begin{figure}[h!]
\centering
\hspace{-0.05in}
    \includegraphics[scale=0.3]{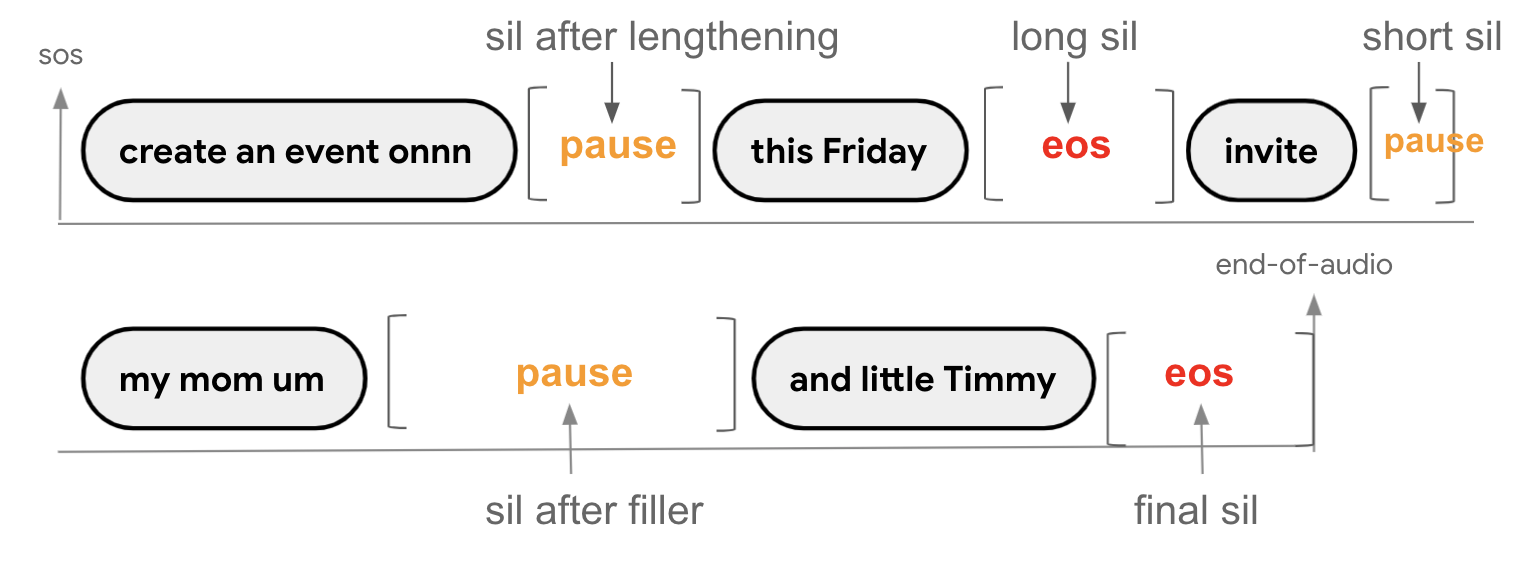}  
    \vspace{-0.1in}
    \caption{An example of annotation}
    \vspace{-0.15in}
    \label{fig:example}
\end{figure}

To model the conversational turn-taking, the first step is to annotate ground-truth disfluencies in the training data. Specifically, we look to label ``user hesitation" as \pause and ``user done speaking" as \eos as an example shown in Fig. \ref{fig:example}. 
%These short-form utterances consists of only single voice queries. Therefore, we  simply append an \eos at the end of utterance while insert \pause for the other silences segments obtained by forced alignment.  
%\TS{i would spell out how you did \pause in more detail. you looked at the forced alignment and if inter-utterance silence is longer then XX seconds you label as pause. but mention your transcript doesnt need an alignment.}
%While short-form utterances could be used to model \eos and regular short pause, it covers limited disfluencies of \pause. In addition, \eos only shows up once at the end of each utterance. The model could incorrectly learn to stop emitting tokens after seeing it, which cause severe problems on longer utterances or utterances with disfluencies. 

Our training data consists of (1) short-form voice search queries from Google’s voice search product, actions on various platforms and (2) long-form data from YouTube, which consists of multiple spontaneous speech sentences including more natural and free voice inputs which include disfluencies. All the utterances are anonymized and hand-transcribed.
%To include disfluencies and long utterances, we explore long-form data from Google’s other speech applications e.g. YouTube. These data includes longer audio consisting of multiple spontaneous speech sentences including more natural and free voice input. 
However, labeling the ground-truth with \pause and \eos is not straightforward as these labels are not annotated. To address the problem, we performs labeling based on acoustic clues and a text-based disfluency detector as summarized in Table \ref{tab:heuristics}. 
%Specifically, we first identify the silence segment by forced alignment and determine the sentence boundaries based on silence duration where short silences are labeled as \pause while long silence represents the end of a complete sentence (\eos). 

Specifically, we first insert an \eos when there is a long silence or at the end of the utterance and a \pause for short silences where the silence segments are obtained by forced alignment.  
%We also append \eos to the end of the last utterance. 
However, disfluenies may also lead to long silence suffix that could be incorrectly labeled as \eos. 

To eliminate common mis-insertions, we relabel the silence suffix of word lengthening, filled pauses and other disfluency words as \pause. 
%where lengthening is identified if phoneme of last utterance in the word has a duration exceeds 10 standard deviation based on the pre-computed phoneme. 
To do this, we approximate word lengthening by looking at phoneme duration. The means and standard deviations of phonemes are computed based on the training set. If the phoneme of the word end exceeds 10 standard deviation away, we mark it as word lengthening and the silence following it as \pause.
For the filled pause, we simply relabel the silence following predefined filler words as \pause. Finally, we exploit the disfluency detector \cite{Johan21} based on small vocabulary BERT to identify reparandum and interregnum (e.g. “you know”, “well”, “I mean”) of disfluencies. The silence following the identified phrases are labeled as \pause. 
%\TS{we should remember to thank Johann in the acks. did we finally end up using their disfluency op btw in the final system?}
%\TS{i'd spend another stencen on this. specifically we approximate measuing word lengthening by looking at phoneme durations. we find the mean/stddev of phonemes in training and if the phoneme of the word end is more then X stddev away, we mark this as eos and label.} 
%An example of the \pause and \eos annotation is shown in Fig. \ref{fig:example}. \TS{i would try to mention this example towards the beginning so people can relate to it as they are reading.}

\begin{table}[]
\centering
\caption{Rules and exceptions for inserting \eos or \pause annotations.}
\label{tab:heuristics}
\resizebox{\columnwidth}{!}{%
\begin{tabular}{llll}
\toprule 
  \begin{tabular}[c]{@{}l@{}}\quad\\ When this happens...\end{tabular} &
  \begin{tabular}[c]{@{}l@{}}\quad\\ It's likely because...\end{tabular} &
  \begin{tabular}[c]{@{}l@{}}Therefore,\\ insert...\end{tabular} \\
\midrule
Long silence between words  & Speaker finished & \texttt{<eos>} \\[1mm]
Silence following last word & Speaker finished & \texttt{<eos>} \\
\midrule
  Short silence between words  & Speaker not finished & \pause \\[1mm]
  Silence following lengthened words & Speaker not finished & \pause \\[1mm]
  Silence following filler words & Speaker not finished & \pause \\[1mm]
  \begin{tabular}[c]{@{}l@{}}Silence following phrases \\ identified by disfulency detector\end{tabular} &
  \begin{tabular}[c]{@{}l@{}}\quad\\ Speaker not finished\end{tabular} &
  \begin{tabular}[c]{@{}l@{}}\quad\\ \pause \end{tabular} \\
  
  %\begin{tabular}[c]{@{}l@{}}Silence following lengthened words\end{tabular} &
  %\begin{tabular}[c]{@{}l@{}}\quad\\ Speaker not finished\end{tabular} &
  %\begin{tabular}[c]{@{}l@{}}\quad\\ \pause \end{tabular} \\[2mm]
  %\begin{tabular}[c]{@{}l@{}}Silence following filler words \end{tabular} &
  %\begin{tabular}[c]{@{}l@{}}\quad\\ Speaker not finished\end{tabular} &
  %\begin{tabular}[c]{@{}l@{}}\quad\\ \pause \end{tabular} \\[2mm]
  %\begin{tabular}[c]{@{}l@{}}Silence following phrases \\ identified by disfulency detector\end{tabular} &
  %\begin{tabular}[c]{@{}l@{}}\quad\\ Speaker not finished\end{tabular} &
  %\begin{tabular}[c]{@{}l@{}}\quad\\ \pause \end{tabular} \\
\bottomrule
\end{tabular}%
}
\end{table}

\section{Models}
\label{sec:system}

%\begin{figure}[!t]
%\centering
%  \begin{subfigure}[b]{\linewidth}
%     \centering
%     \includegraphics[scale=0.3]{figures/encoder_ep2.png}
%     \caption{{Encoder based conversation turn-taking detector.}}
%     \label{fig:system_encoder}
%     \vspace{0.1in}
%  \end{subfigure}
%  \hfill
%  \begin{subfigure}[b]{\linewidth}
%     \centering
%     \includegraphics[scale=0.3]{figures/decoder_ep2.png}
%     \caption{{Semantic based conversation turn-taking detector.}}
%     \label{fig:system_decoder}
%  \end{subfigure}
%  \vspace{-0.25in}
%   \hfill
%  \begin{subfigure}[c]{\linewidth}
%     \centering
%     \includegraphics[width=\linewidth]{figures/e2e_ep2.png}
%     \caption{{E2E conversation joint layer.}}
%     \label{fig:system_e2e}
%  \end{subfigure}
%   \vspace{0.2in}
%   \hfill
%  \caption{A comparison between the proposed models.}
%  \label{fig:system}
%  \vspace{-0.25in}
%\end{figure}

\begin{figure}[h!]
 \centering
% %begin{center}
 \begin{minipage}[b]{0.25\textwidth}
 \centering
 \includegraphics[width=1.0\linewidth]{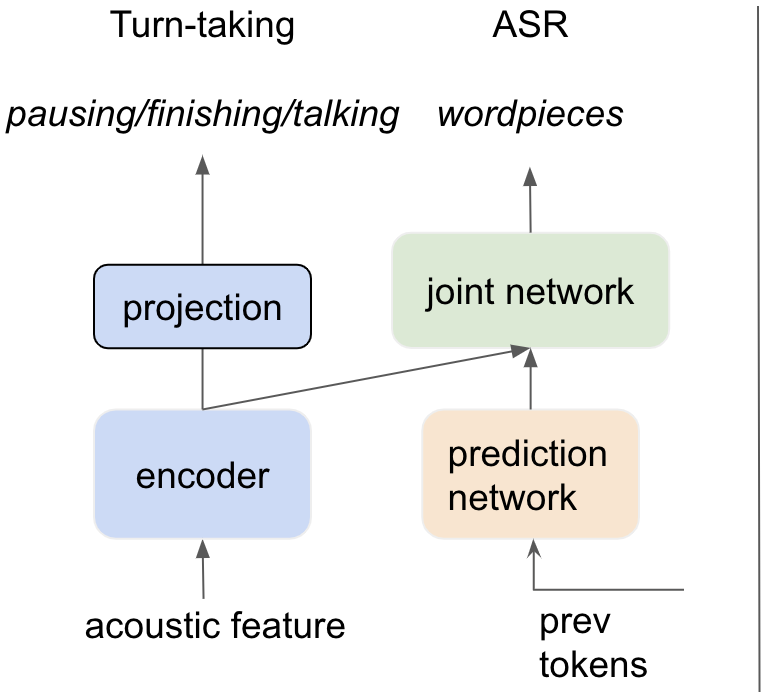}
 \caption{{\it Acoustic based turn-\\taking detector.}} \label{fig:system_encoder}
 \end{minipage}%
 \begin{minipage}[b]{0.21\textwidth}
 \centering
 \includegraphics[width=1.0\linewidth]{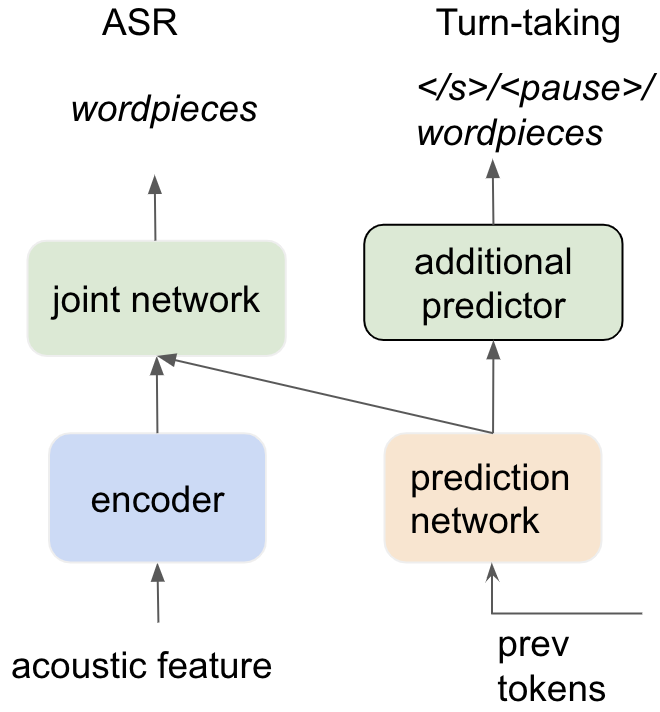}
 \caption{{\it Text based turn-taking predictor.}} \label{fig:system_decoder}
 \end{minipage}
 %\caption{A comparison between the proposed models.}
 %\label{fig:fig:system_base}
 %\end{center}
\end{figure}

%\begin{figure}[h!]
%\centering
%\hspace{-0.05in}
%    \includegraphics[scale=0.31]{figures/enc_dec_ep.png}
%    \vspace{-0.1in}
%    \caption{baseline systems: (a) acoustic detector and (b) text detector}
%    \vspace{-0.15in}
%    \label{fig:system_base}
%\end{figure}

\begin{figure}[h!]
\centering
\hspace{-0.05in}
    \includegraphics[scale=0.31]{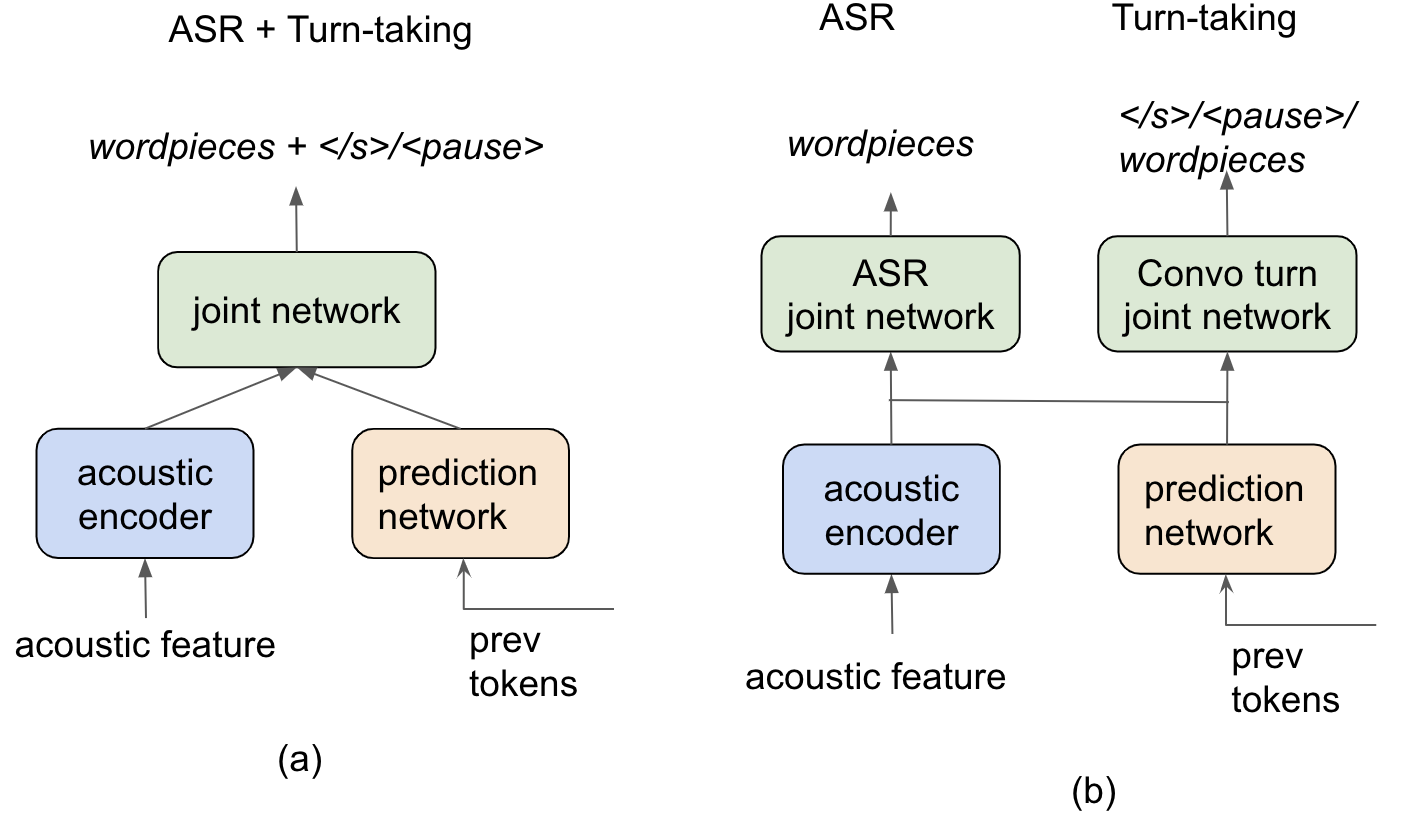}
    \vspace{-0.1in}
    \caption{(a) E2E detector (b) E2E detector with conversation joint network}
    \vspace{-0.15in}
    \label{fig:system_e2e}
\end{figure}

After finishing expanding the transcript with \eos and \pause as described in \ref{sec:annonation}, we look to train the E2E model with these extra tokens to do both ASR and turn-taking in this Section.
To model conversation turn-taking, we explore models that share different components of an E2E Recurrent Neural Network Transducer (RNN-T) \cite{graves2012sequence} based ASR system. RNN-T consists of an encoder, a prediction network and a joint layer. The encoder consists of multiple Conformer layers \cite{gulati2020conformer}. The prediction network summarizes a history of previous predictions into a hidden representation. The joint layer then combines the encoder and the prediction network outputs to predict a wordpiece token given the speech inputs. In the following section, we describe different approaches to build the turn-taking detector, specifically on top of the encoder, prediction network and joint layer respectively.

%\TS{perhaps you should mention here that after relabeling the transcript with EOS and pause, in this section you look to train the E2E model with these extra tokens to do both ASR and turn-taking}
\subsection{Acoustic Detector}

%\TS{so from annotation you get a change in transcript but now you are building something frame level. perhaps you should more clearly descibe in the previous section that once you reannotate you have something frame level, and then depending on if you need per frame or per transcript you can clarify that here.}
 
In this section, we build a frame-level turn-taking detector based on acoustic observations as illustrated in Figure~\ref{fig:system_encoder}. 
To build the architecture, we simply add a projection layer on top of the encoders to output the turn-taking targets, talking, pausing (\texttt{<pause>}) and finishing (\texttt{</s>}). As the encoders are shared, the architecture can achieve a better synchronization between the ASR model and the turn-taking detector, which is important to ensure correct interaction for natural conversational input. 
%In addition, it leverage the existing layers in the ASR encoder without creating another large model. 

We convert all the wordpiece symbols to the talking class and obtain the frame-level targets based on silence alignment.
For each input speech frame $\mathbf{x}_t$ at timestamp \textit{t}, the encoder detector computes the probability distribution of talking, pausing and finishing based on observations $\mathbf{x}_t,... \mathbf{x}_{t-k}$ where \textit{k} represents total context window received by the conformer encoder. We can express the probability of \pause and \eos as:
\begin{equation}
	P(\pause|\vx_{t}, \cdots, \vx_{t-k}), \;
    P(\eos|\vx_{t}, \cdots, \vx_{t-k}) 
	\label{eqn:encoder}
\end{equation}
where the probabilities are thresholded to obtain the pausing or finishing decision. 

%We share the encoders between ASR and turn-taking prediction. By sharing the encoders, the architecture can achieve a better synchronization between the ASR model and the turn-taking detector, which is important to ensure correct interaction for natural conversational input. In addition, it leverage the existing layers in the ASR encoder without creating another large model. 
%The turn-taking task is sensitive to latency as late prediction can introduce a slow response. 
%To avoid that sequence-to-sequence trained E2E ASR models delay token predictions, a simple yet effective technique, FastEmit \cite{yu2021fastemit}, is adopted to address this problem. 

\subsection{Text Predictor} \label{sec:text_pred}
%\TS{i'd move this after the encoder turn-take to be consistent with the figure.}
In this section, we investigate a turn-taking detector that is purely driven by the wordpiece sequence history while discarding the acoustic clues. 
Figure~\ref{fig:system_decoder} illustrates the additional predictor consisting of a fully connected network built on top of the prediction network. As the acoustic observations are skipped, the text based detector is basically a language model predicting probability of the next word being \pause or \eos given the sequence of word pieces already present. 
%To build the semantic based detector, we  simply remove the connection from encoder to the conversation turn-taking joint layer. Hence, it can only receive inputs from the prediction network. 

To train the model, we first optimize the encoder, prediction network and joint network to predict conventional wordpiece label sequences. 
%The prediction network provides a representation conditioned on the $N$ previous (non-blank) labels $\{y_{u-1}, y_{u-2}, ..., y_{u-N+1}\}$ \cite{Rami21}. 
Next, we train the additional predictor with all the other parameters frozen. 
The additional predictor receives the prediction network output conditioned on $N$ previous (non-blank) label predictions \cite{Rami21}
%$\{y_{u-1}, y_{u-2}, ..., y_{u-N+1}\}$ \cite{Rami21} 
and directly output the label sequences including wordpieces as well as \pause and \eos tokens. Thus, the additional predictor is essentially a next word predictor running over the space of wordpieces plus \pause and \texttt{</s>}. 

During inference, we pass top hypothesis to the additional predictor to compute a probability distribution over expanded labels while only take the posterior of \pause and \eos for turn-taking decisions, which could be expressed as:
\begin{equation}
	P(\pause|y_{u-N}, \ldots, y_u),
	P(\eos|y_{u-N}, \ldots, y_u)
	\label{eqn:decoder}
\end{equation}
%Eq. \ref{eqn:decoder} skips the acoustic observations compared to Eq. \ref{eqn:joint}.

\subsection{E2E Detectors} \label{sec:e2e_joint}
 
%The acoustic detector leverages acoustic features without explicit word sequence history while text predictor missed. 
Both acoustic clues and spoken words provide useful features for turn-taking decisions. To exploit both features, a straightforward approach is to build an E2E RNN-T optimizing for wordpieces, \pause and \eos prediction as illustrated in Figure \ref{fig:system_e2e}a, referred as E2E detector. At each time step $t$, the model receives a new acoustic frame $x_t$ and outputs a probability distribution over $y_t \subset \{V \cup \pause \cup \eos\}$, V being the wordpiece vocabulary and a blank symbol.

However, the E2E model could degrade ASR quality as the \pause and \eos tokens do not provide informative features for wordpiece prediction, thus decrease the effective context window instead. %In our experiments, we observed 6\% relative WER increase based on our experiments. 
To ensure that recognition quality is consistent with the conventional ASR, we adapt the model architecture by introducing separate joint networks for ASR joint network and conversation turn joint network as illustrated in Figure~\ref{fig:system_e2e}b, referred as E2E additional joint. The conversation turn joint network is responsible for the turn-taking decisions while the ASR joint layer decodes the wordpieces. 

We perform two stages training strategy similar to text predictor in Sec. \ref{sec:text_pred}. Specifically, we first optimize the encoder, prediction network and the ASR joint layer to predict wordpiece label sequence. Next, we initialize the conversation turn joint network with the ASR joint network. The conversation turn joint network is then fine-tuned with the expanded label sequence including wordpieces, \pause and \eos to adapt the parameters with respect to the additional loss due to extra tokens insertion.
Thus, the conversation turn joint network is able to predict distributions of \pause and \eos given the existing encoder and prediction network outputs.
During inference, we rely on the ASR joint network for beam search decoding over wordpiece space:
\begin{equation}
y* = \operatorname*{arg\,max}_y \log P_{asr}(y|\vx_{t-k}, \cdots, \vx_{t}, y_{u-N}, \ldots, y_u)
	\label{eqn:beam-search}
\end{equation}
%where $\mathbf{y}_0,... \mathbf{y}_{u(t-1)}$ stands for the word-piece sequences emitted thus far. 
At each time step, the conversation turn joint network computes the the probability of the \pause and \eos given the decoding paths obtained by ASR joint network using Eq. \ref{eqn:beam-search}. and acoustic observations. The posterior of \pause and \eos can be expressed as: 
\begin{equation}
    \begin{aligned}
	P_{convo}(\pause|\vx_{t-k}, \cdots, \vx_{t}, y_{u-N}, \ldots, y_u), \\
	P_{convo}(\eos|\vx_{t-k}, \cdots, \vx_{t}, y_{u-N}, \ldots, y_u)
	\end{aligned}
	\label{eqn:joint}
\end{equation}
%An turn-taking decisions is declared if the corresponding probability exceeds a predefined threshold. 
Unlike Eq. \ref{eqn:encoder} or Eq. \ref{eqn:decoder} that predicts decisions based on either the acoustic observations or the word sequences alone, Eq. \ref{eqn:joint} learns the turn-taking objectives from both. The sequence-to-sequence training is performed without alignment, thus we use the FastEmit \cite{yu2021fastemit} regularization to encourage paths that outputs tokens earlier.
%\TS{here you should clarify you train on the annotated transcripts without any alignment. to get the model to emit early we use fast emit.}

\section{Experimental Setup}
\label{sec:exp}

\subsection{Data}

The training data covers utterances from short-form voice query and long-form voice transcription data. The short-form voice query data consists of around 15M utterances collected from Google’s voice search product and actions on various platforms while long-form voice transcription includes 10M of utterances obtained from YouTube. The utterances are anonymized and hand-transcribed.
%and the pausing and finishing labels are added as described in Section \ref{sec:annonation} 
%where forced alignment is based on an existing LSTM-based recognizer to generate silence regions.
%The training data is a combination of clean utterances, simulated noisy utterances with background speech at an average SNR of 13 dB and simulated utterances with other noises at an average SNR of 16dB \cite{Chanwoo17}.
 In  addition  to  the diverse training sets, multi-condition training (MTR) \cite{Chanwoo17} are also used to further increase data diversity.
 %random data down-sampling to 8kHz and SpecAug \cite{park2020specaugment} are also used to further increase data diversity.

To create an evaluation set with disfluency, we first design the scripts based on common voice queries.
%including (1) creating a reminder, event or family note, (2) sending, replying to a broadcast, (3) setting a timer or alarm and (4) making a call. 
Each script contains multiple continued queries as an example shown in Fig. \ref{fig:example}. 
%e.g. "\textit{Put a note in my calendar. Lets call it weekly swim practice. Set it for 5 p.m.}." 
For each script, the speakers insert 4 different types of disfluency including random pauses, filled pauses, word lengthening and repeating phrase. Finally, the speakers manually annotate pausing or finishing labels and the corresponding timestamps. Totally, the evaluation set consists of 200 utterances recorded by 10 speakers, called natural conversation set. We also include 14K voice queries from anonymized Google's voice search set to ensure the recognition quality on a large set.
%The goal of evaluating on typical voice queries is to make sure the models don't degrade performance on a large set.
%\TS{mention the goal when evaluating VS is to make sure natcon still works for general queries and does not degrade.}
%\begin{table}[h!]
%    \centering
%    \caption{Type of disfluency}
%    \begin{tabular}{|c|c|} \hline
%    Type & Example \\ \hline
%    random pauses between words & - \\ \hline
%    inserting filler words & um, hmm, mmm  \\ \hline
%    word lengthening & weatherrrrr \\ \hline
%    repeating phrase & on on \\ \hline   
%    \end{tabular}
%    \label{tab:dis_type}
%\end{table}

% test set stats: http://shortn/_khwVeBCNYA

\subsection{RNN-T Model Architecture}

The RNN-T models use 128D log-Mel features. 
%These features are stacked with 3 frames to the left and downsampled to a 30ms frame rate. SpecAugment \cite{park2020specaugment} is used to improve models' robustness against noise where two frequency masks with a maximum length of 27 and two time masks with a maximum length of 50 are used. 
The encoder network architecture consists of 12 Conformer layers where each layer is of 512 dimension following \cite{sainath2021efficient}. The Conformer layers consist of causal convolution and left-context attention layers where 8-head attention is used in the self-attention layer and the convolution kernel size used is 15. The RNN-T decoder consists of a prediction network and a joint network with a single feed-forward layer with 640 units. The embedding prediction network \cite{Rami21} uses an embedding dimension of 320, and has 1.96M parameters. E2E models are trained to predict 4,096 word pieces~\cite{Schuster2012} plus \pause and \eos. 
%The final model has 140M parameters. It is optimized by minimizing the RNN-T loss. 
We also add the FastEmit \cite{yu2021fastemit} regularization with a weight of 5e-3 to improve the model's prediction latency.  There is no 2nd-pass used for these experiments.

%All the models are trained in Tensorflow \cite{AbadiAgarwalBarhamEtAl15} on 8 × 8 Google's  Tensor Processing Units (TPU) slices with a global batch size of 4,096 using the Lingvo \cite{shen2019lingvo}. Models are trained with 512 TPU cores. Models are optimized using synchronized stochastic gradient descent. We use the Adam optimizer \cite{KingmaBa15} with parameters $\beta_1$=0.9 and $\beta_2$=0.999. A transformer learning rate schedule \cite{Vaswani17} with peak learning rate 1.8e-3 and 32K warm-up steps is used. Exponential moving average has been used to stabilize the model weight updates.
\subsection{Evaluation metrics}
%\TS{i would move. evaluation metric into its own section in experiments, not results}
To evaluate the quality, of detectors we compute the recall and precision rate for both pausing and finishing. 
%To achieve this goal, we compare the ground-truth with the hypothesized ones where the timestamp of true labels are obtained by forced alignment and then manually fix if the alignment is off due to the disfluencies. 
%We define a true pausing or finishing label as detected only if it is detected within 1 sec latency to avoid accepting hypothesized boundaries. 
%When there is more than one hypothesized pausing or finishing belong to the window, we select the one where the corresponding transcription aligned best to the ground-truth. 
%We use edit distance to align the hypothesized pausing and finishing boundaries with the true boundaries.
%A true pausing or finishing label is paired with the hypothesized pausing or finishing 
We first align the hypothesized transcription and pausing/finishing to the reference to pair the predictions with true labels.
%This align process pairs the reference and the hypothesized \pause and \eos, hence we could easily calculate the precision, recall and latency. 
%The deletions of pausing or finishing degrade the recall rates while the insertions hurt the precision rates.
Then, we could easily calculate the precision, recall and latency based on the alignment.
A good recall rate of \eos is critical to ensure the queries are detected and responded.  
The precision rate of \eos affects if the system could interrupt user as false emission causes low precision rate.
To evaluate latency, we measure the time difference between the system predicted pausing or finishing and the true labels. We only calculate the latency from the detected pausing or finishing.
We also evaluate the WERs while only to ensure that adding the turn-taking detectors doesn't hurt recognition quality. The WERs of ASR without turn-taking detectors are  6.3\% for the 14k voice search test set and 10.1 \% for the natural conversation test set. 
%\TS{how do you make sure. also I don't see any WER number reported in Table 3 or 4. If you are sweeping eos and pause thresholds its hard to optimize both. In your eng review you had a table on WER, would it not make sense to add that in here too?}. \SY{eos and pause here is just like a prefetch, which don't trigger any action that affect WER, so it's fixed. There is only one experiment having WER degradation, which is the E2E "without" joint layer but I don't report that one here, so that's why the WER is fixed.}

%\begin{table}[h!]
%    \centering
%    \caption{Metrics}
%    \begin{tabular}{|c|c|} \hline
%    Metrics & Possible impact if it's low \\ \hline
%   Recall of \eos & missing response to user \\ \hline
%    Precision of \eos & interrupt user  \\ \hline
%    Recall of \pause & less responsive  \\ \hline
%    Precision of \pause & redundant back-channeling   \\ \hline
%    \end{tabular}
%    \label{tab:metrics}
%\end{table}

%\begin{table} [h!]
%\caption{WER for all systems}
%\centering
%\begin{tabular}{|c|c|c|} \hline
%	    & VS    & NatCon  \\ \hline
%	WER & 6.3\% & 10.1\%  \\ \hline
%\end{tabular}
%\label{table:wers}
%\end{table}

\section{Results}

In this section we present the experimental results comparing the turn-taking detectors as described in \ref{sec:system}. 
We first investigate the WERs of each model on typical voice-search set and natural conversation set. Table \ref{table:wer} demonstrates that E2E detector could increase the WERs by 6\% relatively, which implies that directly optimizing the conventional RNN-T to both tasks would hurt recognition quality as suggested in Section \ref{sec:e2e_joint} while the recognition quality could be fixed by introducing additional joint network i.e. E2E add. joint in Table \ref{table:wer}. 
Hence, we only compare the acoustic detector, text predictor and E2E with additional joint for the following experiments on detection accuracy and latency.

\begin{table} [h!]
\caption{WER of each model. VS is the typical voice-search set; Convo is the natural conversation set.}
\centering
\vspace{-0.05in}
\begin{tabular}{|c|c|c|c|c|} \hline
	WER &  Acoustic & Text & E2E & E2E  \\ 
	 (\%)   &  detector & predictor & detector & add. joint  \\\hline
	VS set & 6.3 & 6.3 & 6.7 & 6.3\\ \hline
	Convo set & 10.1 & 10.1 & 10.6 & 10.1 \\\hline
\end{tabular}
\label{table:wer}
\end{table}

\begin{figure}[h!]
\hspace{-0.05in}
    \includegraphics[scale=0.39]{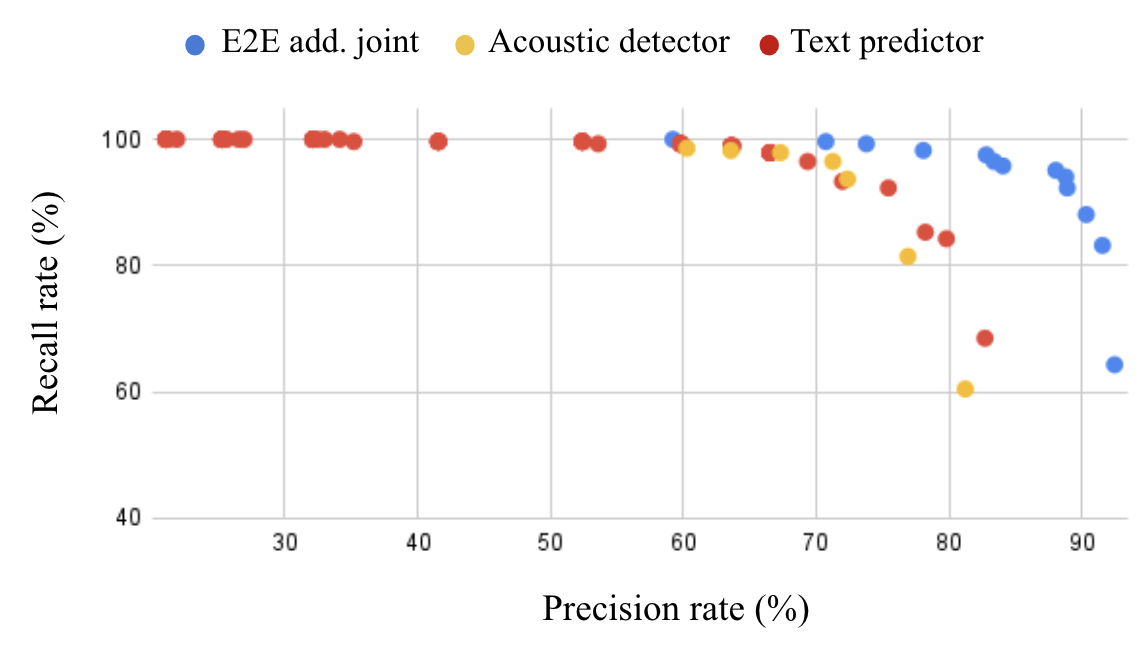}
    \vspace{-0.1in}
    \caption{Precision-Recall curve of finishing speaking.}
    %\vspace{-0.15in}
    \label{fig:pr_eos}
\end{figure}

Figure \ref{fig:pr_eos} reports the PR (Precision-Recall) curves for both \eos and \pause. Upper curves are better. The curve is obtained by sweeping the thresholds of posterior of \eos and \pause obtained by Eq. \ref{eqn:encoder}, \ref{eqn:decoder} and \ref{eqn:joint}.
Figure \ref{fig:pr_eos} demonstrates that the system based on the E2E with additional joint is better than acoustic and text based detector, both of which degrade rapidly for the region where precision rate is over 70\%. Table \ref{table:pr_eos} is the optimal operating point obtained in Figure \ref{fig:pr_eos}. Table \ref{table:pr_eos} shows that although both the acoustic and text detector could achieve a good recall rate covering over 97\% true \eos with a small median latency of around 100 ms, both systems have a precision rate below 70\% which indicates that pausing are misclassified as finishing.
The problem has been largely improved using the E2E with additional joint where precision rate has been improved by roughly 18\%. The results reveal that while either acoustic observations or word sequences alone could easily identify pausing of disfluency as finishing, and combined modeling can significantly remedy the problem.  

\begin{figure}[h!]
\hspace{-0.05in}
    \includegraphics[scale=0.39]{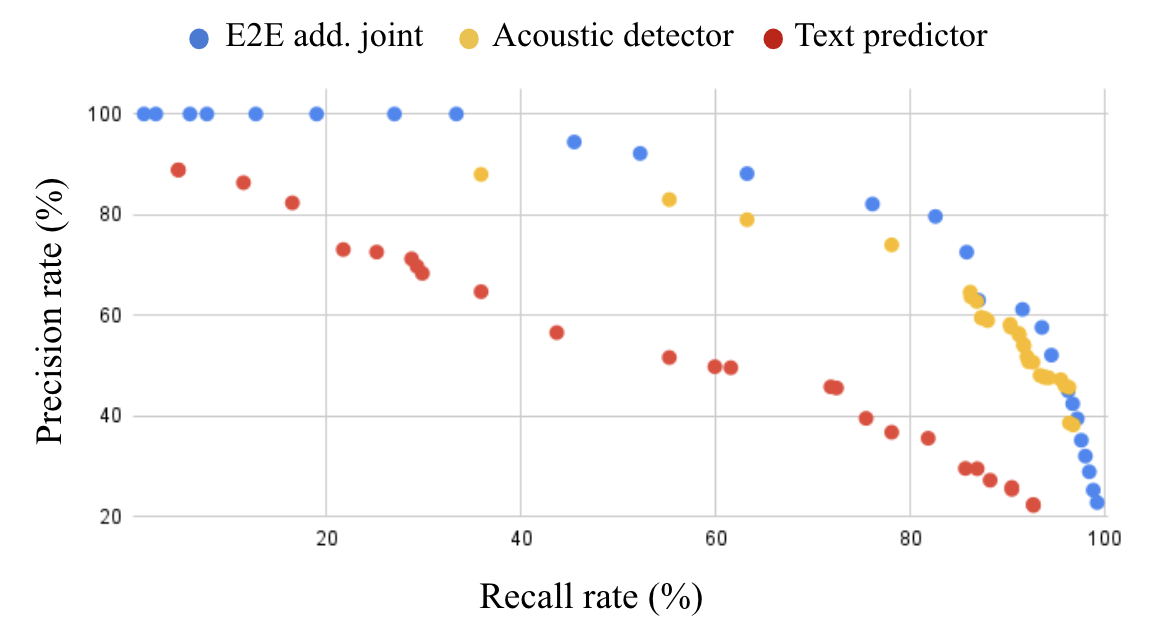}
    \vspace{-0.1in}
    \caption{Precision-Recall curve of pausing.}
    %\vspace{-0.15in}
    \label{fig:pr_pause}
\end{figure}

In Figure \ref{fig:pr_pause}, we compare the 3 systems for pausing detection. Figure \ref{fig:pr_pause} reveals that text predictor is much worse than the other two systems on predicting pausing. This indicates that it is difficult to predict pausing based on only word sequences history. The system based on E2E with additional joint still performance best over the 3 systems where both recall and precision are over 10\% better than the acoustic detector as shown in Table \ref{table:pr_pause}.

\begin{table} [h!]
\caption{Precision, recall and latency for finishing speaking.}
\centering
\vspace{-0.05in}
\begin{tabular}{|c|c|c|c|c|} \hline
	model &  recall & precision & 50th & 90th  \\
	      &    (\%)     &      (\%)     & latency & latency  \\ \hline
	Acoustic detector & 97.3 & 67.3 & 90 ms & 330 ms\\ \hline
	Text predictor &  97.2 & 66.5 & 90 ms & 200 ms\\ \hline
	E2E add. joint & 97.5 & 84.7 & 100 ms & 240 ms \\ \hline
\end{tabular}
\label{table:pr_eos}
%\vspace{-0.15in}
\end{table}
%\vspace{-0.05in}

\begin{table} [h!]
\caption{Precision, recall and latency for pausing.}
\centering
\vspace{-0.05in}
\begin{tabular}{|c|c|c|c|c|} \hline
    model &  recall & precision & 50th & 90th  \\
	      &    (\%)     &      (\%)     & latency & latency  \\ \hline
	Acoustic detector &  72.5 & 60.0 & 270 ms & 790 ms\\ \hline
	Text predictor & 29.8 & 69.7 & 70 ms & 930 ms\\ \hline
	E2E add. joint & 84.8 & 77.5 & 300 ms & 840 ms \\ \hline
\end{tabular}
\label{table:pr_pause}
%\vspace{-0.15in}
\end{table}
%\vspace{-0.1in}

%\subsection{Analysis}

%TODO(shuoyiin): put some examples here

\section{Conclusion}
In this work, we incorporate a turn-taking detector into an unified E2E RNN-Transducer by sharing the encoder and the prediction network while adapting the joint layers to optimize both the recognition and turn-taking detection. The proposed approach demonstrates over 97\% recall rate and 85\% precision rate on predicting true turn-taking with only 100 ms latency on difficult continued queries with 4 types of disfluencies. The analyses reveal that pure acoustic or text based predictor achieve comparable performance on detecting finishing while acoustic observations are much more useful for pausing detection. 
%The combined modeling can significantly improve the detection accuracy compared with the systems using  the problem

\section{Acknowledgement}
We would like to thank to Jon Bloom and Jaclyn Konzelmann for natural conversation set design
and Johann Rocholl and Dan Walker for providing the disfluency detector for labeling.

\newpage
\bibliographystyle{IEEEbib}
\bibliography{main}

\begin{thebibliography}{10}

\bibitem{ibm2015}
S.~Thomas, G.~Saon, M.~V. Segbroeck, and S.~Naranyanan,
\newblock ``Improvements to the ibm speech activity detection system for the
  darpa rats program,''
\newblock {\em ICASSP}, 2015.

\bibitem{chang2019joint}
Shuo-Yiin Chang, Rohit Prabhavalkar, Yanzhang He, Tara~N Sainath, and Gabor
  Simko,
\newblock ``Joint endpointing and decoding with end-to-end models,''
\newblock in {\em ICASSP 2019-2019 IEEE International Conference on Acoustics,
  Speech and Signal Processing (ICASSP)}. IEEE, 2019, pp. 5626--5630.

\bibitem{chang2019unified}
Shuo-Yiin Chang, Bo~Li, and Gabor Simko,
\newblock ``A unified endpointer using multitask and multidomain training,''
\newblock in {\em 2019 IEEE Automatic Speech Recognition and Understanding
  Workshop (ASRU)}. IEEE, 2019, pp. 100--106.

\bibitem{chang2017endpoint}
Shuo-Yiin Chang, Bo~Li, Tara~N Sainath, Gabor Simko, and Carolina Parada,
\newblock ``Endpoint detection using grid long short-term memory networks for
  streaming speech recognition.,''
\newblock in {\em Interspeech}, 2017, pp. 3812--3816.

\bibitem{li2020towards}
Bo~Li, Shuo-yiin Chang, Tara~N Sainath, Ruoming Pang, Yanzhang He, Trevor
  Strohman, and Yonghui Wu,
\newblock ``Towards fast and accurate streaming end-to-end asr,''
\newblock in {\em ICASSP 2020-2020 IEEE International Conference on Acoustics,
  Speech and Signal Processing (ICASSP)}. IEEE, 2020, pp. 6069--6073.

\bibitem{amazon2018}
R.~Maas, A.~Rastrow, C.~Ma, G.~Lan, K.~Goehner, G.~Tiwari, S.~Joseph, and
  B.~Hoffmeister,
\newblock ``Combining acoustic embeddings and decoding features for
  end-of-utterance detection in real-time far-field speech recognition
  systems,''
\newblock {\em ICASSP}, 2018.

\bibitem{icml18}
D.~Lala, K.~Inoue, and T.~Kawahara,
\newblock ``Evaluation of real-time deep learning turn-taking models for
  multiple dialogue scenarios,''
\newblock {\em ICML}, 2018.

\bibitem{Osakais17}
C.~Liu, C.~Ishi, and H.~Ishiguro,
\newblock ``Turn-taking estimation model based on joint embedding of lexical
  and prosodic contents,''
\newblock {\em Interspeech}, 2017.

\bibitem{EOTis17}
Julian~Hough Angelika Maier~and and David Schlangen,
\newblock ``Towards deep end-of-turn prediction for situated spoken dialogue
  systems,''
\newblock {\em Interspeech}, 2017.

\bibitem{kyoto18}
R.~Masumura, T.~Asami, H.~Masataki, R.~Ishii, and R.~Higashinaka,
\newblock ``Prediction of turn-taking using multitask learning with prediction
  of backchannels and fillers,''
\newblock {\em Interspeech}, 2018.

\bibitem{filler19}
D.~Lala, S.~Nakamura, and T.~Kawahara,
\newblock ``Analysis of effect and timing of fillers in natural turn-taking,''
\newblock {\em Interspeech}, 2019.

\bibitem{li2020comparison}
J.~Li, Y.~Wu, Y.~Gaur, et~al.,
\newblock ``{On the Comparison of Popular End-to-End Models for Large Scale
  Speech Recognition},''
\newblock in {\em Proc. Interspeech}, 2020.

\bibitem{Ryan19}
Y.~He, T.~N. Sainath, R.~Prabhavalkar, et~al.,
\newblock ``{Streaming End-to-end Speech Recognition For Mobile Devices},''
\newblock in {\em Proc. ICASSP}, 2019.

\bibitem{CC18}
C.-C. Chiu, T.~N. Sainath, Y.~Wu, et~al.,
\newblock ``{State-of-the-art Speech Recognition With Sequence-to-Sequence
  Models},''
\newblock in {\em Proc. ICASSP}, 2018.

\bibitem{JinyuLi2019}
J.~Li, R.~Zhao, H.~Hu, and Y.~Gong,
\newblock ``Improving {RNN} transducer modeling for end-to-end speech
  recognition,''
\newblock in {\em Proc. ASRU}, 2019.

\bibitem{Zeyer2020}
A.~Zeyer, A.~Merboldt, R.~Schl\"{u}ter, and H.~Ney,
\newblock ``A new training pipeline for an improved neural transducer,''
\newblock in {\em Proc. Interspeech}, 2020.

\bibitem{Golan16}
G.~Pundak and T.~N. Sainath,
\newblock ``Lower frame rate neural network acoustic models,''
\newblock in {\em Proc. Interspeech}, 2016.

\bibitem{Johan21}
J.~C. Rocholl, V.~Zayats, D.~D. Walker, N.~B. Murad, A.~Schneider, and D.~J.
  Liebling,
\newblock ``Disfluency detection with unlabeled data and small bert models,''
\newblock {\em Interspeech}, 2021.

\bibitem{graves2012sequence}
Alex Graves,
\newblock ``{Sequence transduction with recurrent neural networks},''
\newblock {\em arXiv preprint arXiv:1211.3711}, 2012.

\bibitem{gulati2020conformer}
Anmol Gulati, James Qin, Chung-Cheng Chiu, Niki Parmar, Yu~Zhang, Jiahui Yu,
  Wei Han, Shibo Wang, Zhengdong Zhang, Yonghui Wu, et~al.,
\newblock ``Conformer: Convolution-augmented transformer for speech
  recognition,''
\newblock {\em arXiv preprint arXiv:2005.08100}, 2020.

\bibitem{Rami21}
R.~Botros and T.N. Sainath,
\newblock ``Tied \& reduced rnn-t decoder,''
\newblock in {\em Proc. Interspeech}, 2021.

\bibitem{yu2021fastemit}
Jiahui Yu, Chung-Cheng Chiu, Bo~Li, Shuo-yiin Chang, Tara~N Sainath, Yanzhang
  He, Arun Narayanan, Wei Han, Anmol Gulati, Yonghui Wu, et~al.,
\newblock ``Fastemit: Low-latency streaming asr with sequence-level emission
  regularization,''
\newblock in {\em Proc. ICASSP}. IEEE, 2021, pp. 6004--6008.

\bibitem{Chanwoo17}
C.~Kim, A.~Misra, K.~Chin, T.~Hughes, A.~Narayanan, T.~N. Sainath, and
  M.~Bacchiani,
\newblock ``Generated of large-scale simulated utterances in virtual rooms to
  train deep-neural networks for far-field speech recognition in google home,''
\newblock in {\em Proc. Interspeech}, 2017.

\bibitem{sainath2021efficient}
Arun~Narayanan Tara N.~Sainath, Yanzhang~He et~al.,
\newblock ``An efficient streaming non-recurrent on-device end-to-end model
  with improvements to rare-word modeling,''
\newblock 2021.

\bibitem{Schuster2012}
M.~Schuster and K.~Nakajima,
\newblock ``Japanese and {K}orean voice search,''
\newblock in {\em Proc. ICASSP}, 2012.

\end{thebibliography}
\end{document}